\documentclass{article}

\usepackage[final]{neurips_2023} 

\usepackage{graphicx}
\usepackage{amsmath, amssymb}
\usepackage{booktabs}
\usepackage{microtype}
\usepackage{hyperref}
\usepackage{xcolor}

\usepackage{caption}
\usepackage{subcaption}

\usepackage{algorithm}
\usepackage{algpseudocode}

\title{Federated Learning Under Temporal Drift: \\
Mitigating Catastrophic Forgetting via Experience Replay}

\author{
  Sahasra Kokkula \\
  Columbia University \\
  \texttt{sk5652@columbia.edu}
  \And
  Daniel David \\
  Columbia University \\
  \texttt{dd3269@columbia.edu}
  \And
  Aaditya Baruah \\
  Columbia University \\
  \texttt{abb2237@columbia.edu}
}

\begin{document}
\maketitle

\section{Introduction}
Federated Learning (FL) is appealing because it lets multiple clients collaboratively train a shared model without sending raw data to a central server. In theory, this is ideal for privacy-sensitive settings. In practice though, client data is rarely stable, it changes over time. People's behavior shifts, trends come and go, and the distribution you trained on last week may not look like the distribution you see next month. This kind of non-stationarity (concept drift) is where standard FL pipelines start to struggle, because the model keeps "chasing" the newest data and gradually loses what it learned before.

In this project, we focus on one core question: how do we keep a federated model from collapsing under temporal drift? We simulate a simple but realistic version of drift using Fashion-MNIST: training happens in seasonal phases where only a subset of classes is available in each phase (e.g., winter has coat/boot/pullover, summer has t-shirt/dress/sandal), repeated across 25 communication rounds. Under this setup, standard FedAvg shows severe catastrophic forgetting, performance rises to around 74\% early on, then drops to roughly 28\% once the distribution shifts.

Our main approach is client-side experience replay, a simple but effective continual-learning mechanism integrated into federated training. During local updates, each client maintains a compact memory buffer of past samples and trains on a mixture of buffered data and the current "season" data. We keep the server pipeline unchanged (standard FedAvg aggregation), but the replay mechanism directly addresses the root cause of forgetting by continuously re-exposing the model to previously seen classes. In our experiments, a 50-sample-per-class buffer consistently restores performance to 78-82\%, and our ablation study shows a clear memory-accuracy trade-off as the buffer size increases.

\paragraph{Contributions.}
\begin{itemize}
  \item We empirically demonstrate catastrophic forgetting in FedAvg under a controlled seasonal drift simulation.
  \item We implement and evaluate client-side experience replay as a mitigation strategy, without modifying the server aggregation procedure.
  \item We compare against an IID FedAvg baseline to isolate drift effects from "normal" FL training.
  \item We perform a buffer-size ablation to quantify how much memory is needed to meaningfully reduce forgetting.
\end{itemize}

\section{Related Work}

Federated Learning trains a shared model across many clients without centralizing raw data, and a common baseline is FedAvg, where clients run local SGD and the server periodically averages their updates \cite{mcmahan2017fedavg}. Our setting also fits well with the idea of continuous learning because client data can change over time and models may forget patterns from the past. Rehearsal methods solve this problem by keeping a small memory of past examples and using them in training. Exemplar-based methods like iCaRL show how a small buffer can make stability better when distributions change \cite{rebuffi2017icarl}. We follow the same high-level idea but keep replay entirely client-side inside an FL pipeline.

Non-stationarity is a major stressor in realistic deployments, and prior work links concept drift in federated settings to performance collapse unless it's explicitly handled \cite{casado2021drift}. We use that lens to treat drift as the main challenge and evaluate client-side replay under a controlled drift simulation. Finally, while FL avoids sharing raw data, gradients can still leak information; DLG demonstrates that inputs may be reconstructed from shared gradients in some conditions \cite{zhu2019dlg}, which reinforces our choice to keep replay buffers local. We run experiments on Fashion-MNIST as a standard, simple benchmark for studying these dynamics under shift \cite{xiao2017fashionmnist}.

\section{Methodology}
We run standard FedAvg as our baseline, and then we modify only the client-side training by adding experience replay. Concretely, each client keeps a small local memory buffer of past samples and, at every communication round, trains on a mixture of the current season's data plus the buffered data, then sends only model updates to the server for normal FedAvg averaging. This way, the aggregation logic stays identical, but the local optimization signal no longer "forgets" older classes when they disappear in later seasons; the buffer continuously re-introduces them during training.

To understand how much memory is actually needed (and whether the gains are just coming from "more data"), we also run a buffer-size ablation study, repeating the same FedAvg+replay pipeline with different buffer capacities per class (including 0 as the no-replay control) and comparing final accuracy and stability across rounds to quantify the memory-performance trade-off.

\begin{algorithm}[H]
\caption{Temporal Federated Learning with Client-Side Replay Buffers}
\label{alg:tfl_replay}
\begin{algorithmic}[1]
\Require Seasons $S$, clients $k=1..K$, rounds per season $R$, buffer capacity $B$, local epochs $E$, learning rate $\eta$
\Ensure Final global model weights $w$

\State Initialize global model weights $w$
\For{each client $k$}
    \State Initialize replay buffer $\mathcal{B}_k \leftarrow \emptyset$
\EndFor

\For{each season $s \in S$}
    \For{$r = 1$ to $R$}
        \State Server broadcasts $w$ to participating clients \cite{mcmahan2017fedavg}
        
        \For{each client $k$ (in parallel)}
            \State Load seasonal local data $\mathcal{D}^{(s)}_k$
            \State Sample replay batch $\mathcal{R}_k \subseteq \mathcal{B}_k$ \cite{rebuffi2017icarl}
            \State Train locally for $E$ epochs on $\mathcal{D}^{(s)}_k \cup \mathcal{R}_k$ using SGD($\eta$)
            \State Obtain updated weights $w_k$
            \State Update replay buffer $\mathcal{B}_k$ using samples from $\mathcal{D}^{(s)}_k$ with capacity $B$ \cite{rebuffi2017icarl}
            \State Send $w_k$ (or $\Delta w_k$) to the server
        \EndFor
        
        \State Server aggregates client updates to update $w$ \cite{mcmahan2017fedavg}
        \State Optionally evaluate $w$ on the global test set
    \EndFor
    \State \textit{Note: Seasonal non-stationarity / concept drift setting \cite{casado2021drift}}
    \State \textit{Gradient leakage motivation for keeping buffers local \cite{zhu2019dlg}}
\EndFor

\State \Return $w$
\end{algorithmic}
\end{algorithm}

\section{Experimental Setup}

\subsection{Dataset and Preprocessing}
\begin{itemize}
    \setlength\itemsep{0.3em}
    \item Dataset: Fashion-MNIST (60,000 training, 10,000 test images, 28x28 grayscale)
    \item Classes: T-shirt(0), Trouser(1), Pullover(2), Dress(3), Coat(4), Sandal(5), Shirt(6), Sneaker(7), Bag(8), Ankle-boot(9)
    \item Preprocessing: Pixel values normalized to [0, 1]. No data augmentation applied.
\end{itemize}

\subsection{Client Configuration}
\begin{itemize}
    \setlength\itemsep{0.3em}
    \item Number of Clients: 10 (all participate each round, i.e., full participation)
    \item Data Distribution:
    \begin{itemize}
        \item Phase 0 (Init): 10,000 IID samples split equally (1,000/client)
        \item Seasonal Phases: Non-IID, class-restricted (see drift schedule below)
    \end{itemize}
    \item Client Selection: All 10 clients participate every round (no random sampling)
\end{itemize}

\subsection{Seasonal Drift Schedule}
We simulate temporal concept drift by restricting available classes per season:
\begin{itemize}
    \setlength\itemsep{0.3em}
    \item Phase 0 (Init IID): All 10 classes, balanced - 5 rounds
    \item Phase 1 (Winter): Classes [4-Coat, 9-Ankle boot, 2-Pullover] only - 5 rounds
    \item Phase 2 (Spring): Classes [1-Trouser, 6-Shirt, 8-Bag] only - 5 rounds
    \item Phase 3 (Summer): Classes [0-T-shirt, 3-Dress, 5-Sandal] only - 5 rounds
    \item Phase 4 (Fall): Classes [7-Sneaker, 2-Pullover, 6-Shirt, 1-Trouser] - 5 rounds
\end{itemize}
Total: 25 communication rounds. Each phase transition represents abrupt drift.

\subsection{Model Architecture}
CNN Architecture (same for all experiments):
\begin{itemize}
    \setlength\itemsep{0.3em}
    \item Conv2d (1 $\rightarrow$ 32, 5x5, padding=2) + ReLU + MaxPool(2x2)
    \item Conv2d (32 $\rightarrow$ 64, 5x5, padding=2) + ReLU + MaxPool(2x2)
    \item Flatten $\rightarrow$ Linear (3136 $\rightarrow$ 512) + ReLU $\rightarrow$ Linear (512 $\rightarrow$ 10)
    \item Total Parameters: $\sim$1.7M
\end{itemize}

\subsection{Training Hyperparameters}

Federated Settings:
\begin{itemize}
    \setlength\itemsep{0.3em}
    \item Communication Rounds: 25 total (5 per phase)
    \item Local Epochs: 1 per round
    \item Local Batch Size: 32
    \item Optimizer: SGD with momentum=0.9, lr=0.01
    \item Aggregation: FedAvg (simple weight averaging)
\end{itemize}

Centralized Baseline:
\begin{itemize}
    \setlength\itemsep{0.3em}
    \item Epochs: 10
    \item Batch Size: 64
    \item Optimizer: Adam, lr=0.001
\end{itemize}

\subsection{Experience Replay Configuration}
\begin{description}
    \setlength\itemsep{0.3em}
    \item[Buffer Location:] Client-side (each client maintains own buffer).
    \item[Buffer Update Policy:] Fill-up (add samples until capacity, then stop).
    \item[Buffer Size:] 50 samples per class per client (default).
    \item[Total Buffer Memory:] $50 \times 10 \text{ classes} \times 784 \text{ bytes} \approx 392$KB per client.
    \item[Replay Strategy:] Concatenate buffer with current season data each round.
    \item[Sampling Ratio:] No explicit ratio; all buffered samples used alongside new data.
\end{description}

\subsection{Evaluation Protocol}
\begin{description}
    \setlength\itemsep{0.3em}
    \item[Test Set:] Full Fashion-MNIST test set (10,000 samples, all classes).
    \item[Evaluation Frequency:] After every communication round.
    \item[Metrics:] Global accuracy, per-class accuracy.
    \item[Note:] We evaluate on ALL classes even when training on subset (measures forgetting).
\end{description}

\section{Results and Discussion}

\subsection{Global Accuracy Over Time}
\vspace{-10pt} 

\begin{figure}[H]
    \centering
    \includegraphics[width=0.8\textwidth]{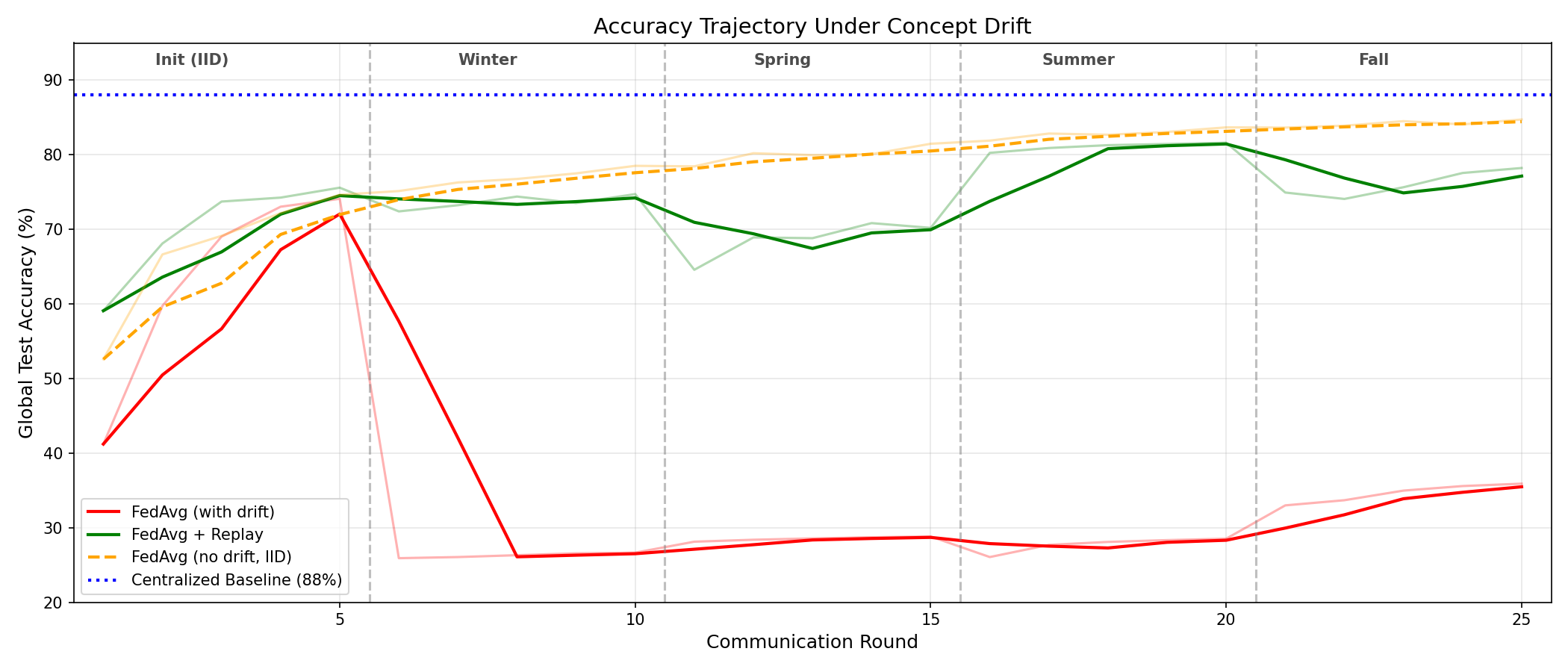}
    \vspace{-10pt} 
    \caption{Global Accuracy Recovery Curve}
    \label{fig:recovery_curve}
    \vspace{-10pt} 
\end{figure}

Key Observations:
\begin{itemize}
    \setlength\itemsep{0em}
    \setlength\parskip{0em}
    \item[] Centralized Baseline: $\sim$88\% (trained on IID data only, upper bound reference)
    \item[] FedAvg on IID (no drift): Maintains $\sim$75-80\% accuracy throughout 25 rounds
    \item[] FedAvg (with drift, Rounds 1-5): Reaches $\sim$74\% accuracy on IID data
    \item[] FedAvg (with drift, Rounds 6+): Drops to 26-36\% as forgetting occurs
    \item[] FedAvg + Replay: Maintains 66-82\% throughout, recovering after each drift
\end{itemize}

The accuracy drop in standard FedAvg from 74\% to $\sim$28\% (a 46 percentage point drop) demonstrates catastrophic forgetting. The IID baseline proves this is due to drift, not FL itself. Experience replay reduces this drop significantly.

\subsection{Per-Class Accuracy Analysis}

\begin{figure}[H]
    \centering
    \includegraphics[width=0.8\textwidth]{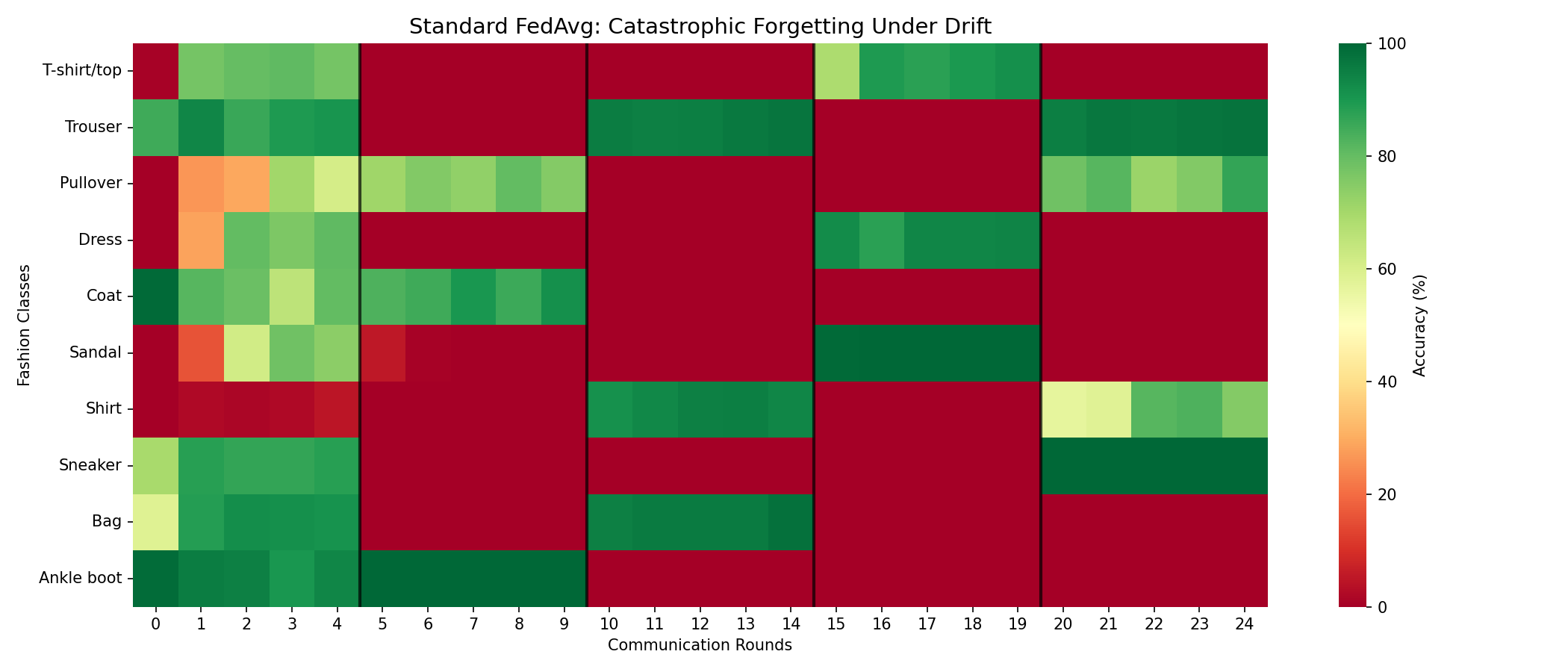}
    \vspace{-10pt}
    \caption{Standard FedAvg: Catastrophic Forgetting Under drift}
    \label{fig:heatmap_fedavg} 
    \vspace{-10pt}
\end{figure}

The heatmap shows complete knowledge loss for classes not in current season. During Summer (Rounds 16-20), only T-shirt, Dress, and Sandal retain accuracy; all other classes drop to 0\%.

\begin{figure}[H]
    \centering
    \includegraphics[width=0.8\textwidth]{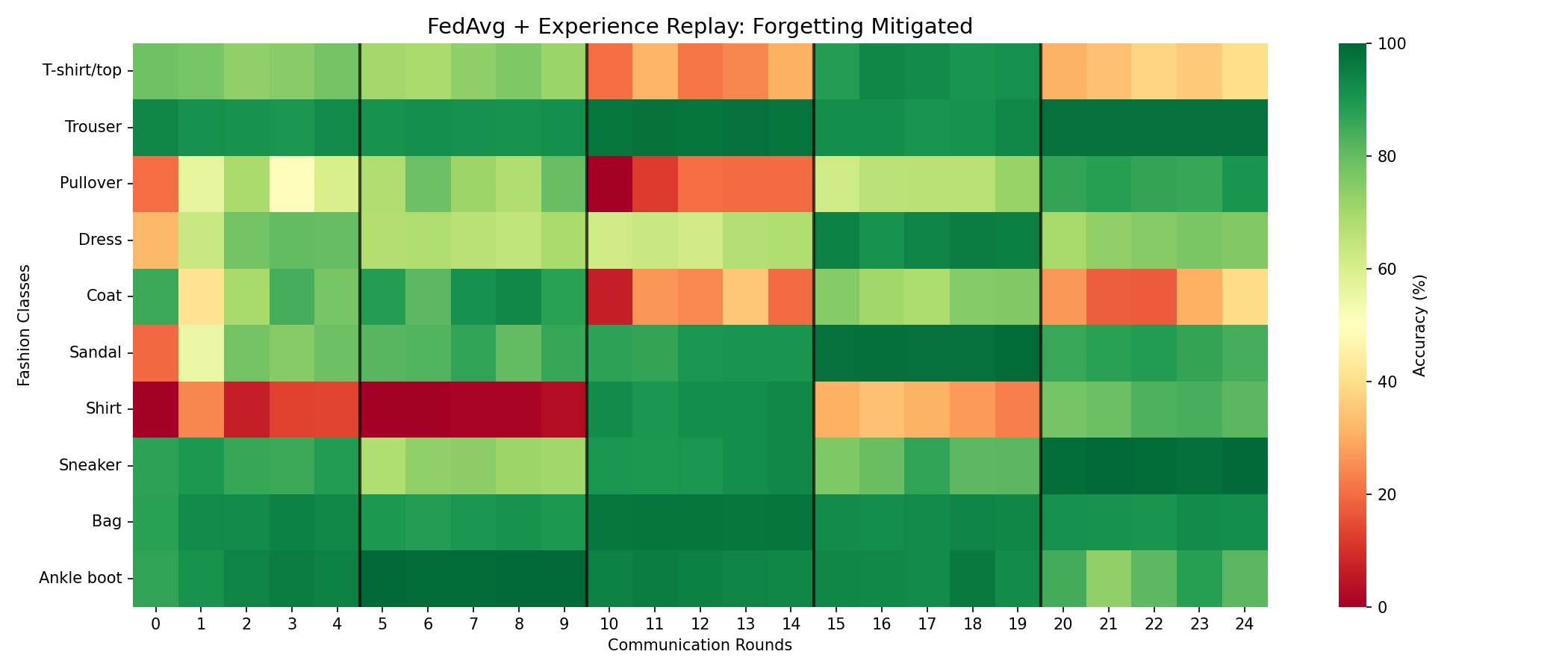}

    \caption{FedAvg + Experience Replay: Forgetting Mitigated}
    \label{fig:heatmap_replay} 
  
\end{figure}

With replay, classes maintain non-zero accuracy across all rounds. While current-season classes show highest performance, buffered samples prevent complete forgetting of other classes. Final round shows balanced performance across most categories.

\subsection{Robustness Profile}
\begin{figure}[H]
    \centering
    \includegraphics[width=0.8\textwidth]{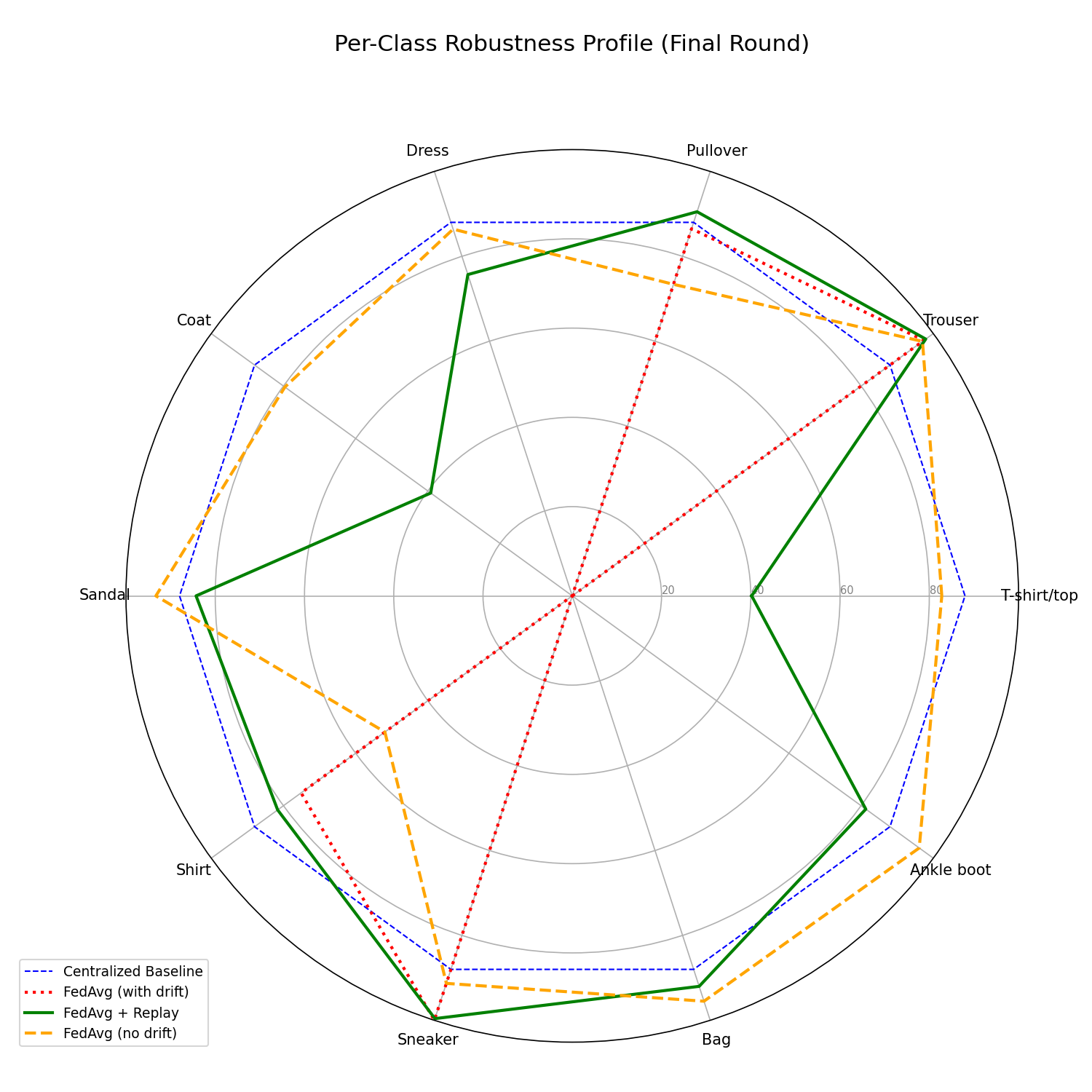}

    \caption{Per class Robustness Profile}
    \label{fig:radar_robustness} 
   
\end{figure}

The radar chart shows final-round per-class accuracy for each approach. FedAvg+Replay (green) achieves more uniform coverage compared to standard FedAvg (red), which shows gaps for forgotten classes. The IID baseline (orange) shows what FL achieves without drift.

\subsection{Buffer Size Ablation}
\vspace{-10pt}

\begin{figure}[H]
    \centering
    \includegraphics[width=0.8\textwidth]{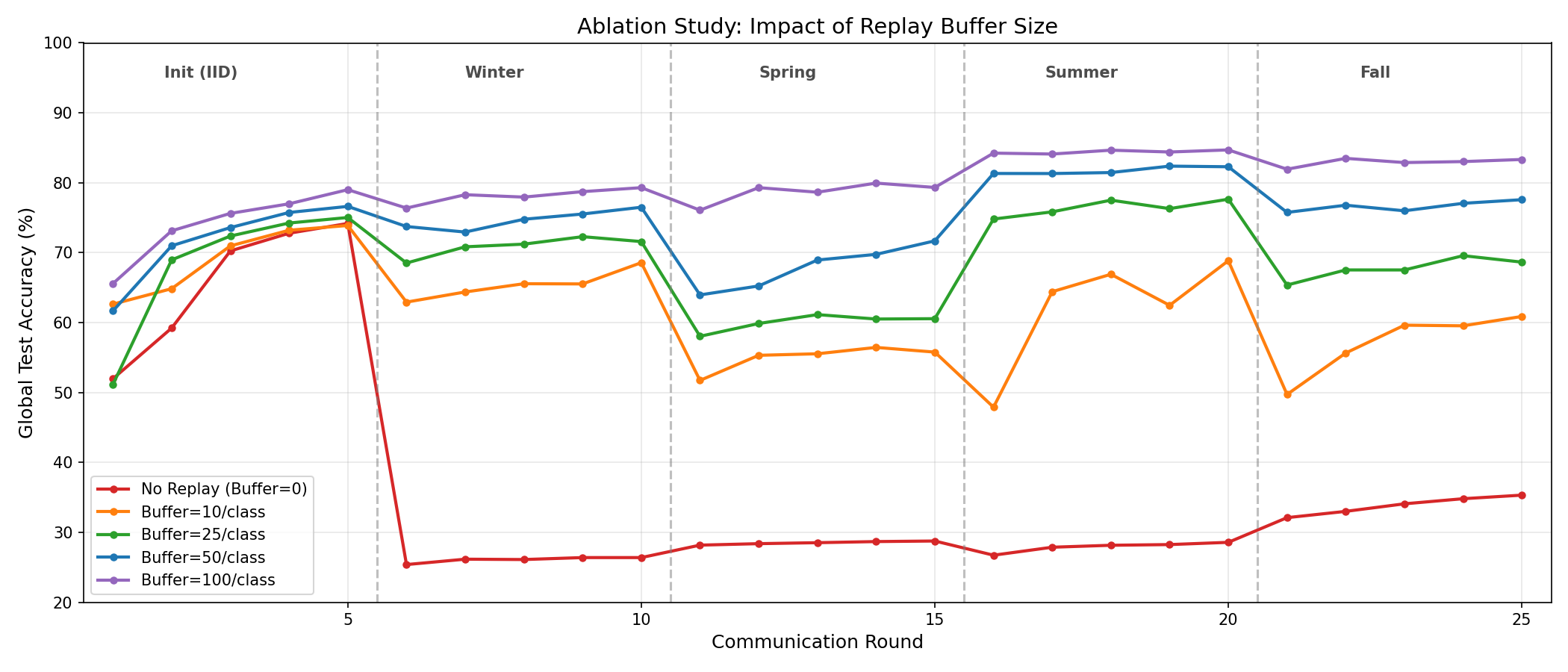}
    \vspace{-10pt}
    \caption{Ablation study: Impact of replay buffer size}
    \label{fig:ablation_curve} 
    \vspace{-10pt}
\end{figure}

We conducted an ablation study varying the replay buffer size:

\begin{table}[H]
\centering
\small
\renewcommand{\arraystretch}{1.15}
\setlength{\tabcolsep}{6pt}
\caption{Ablation Study on Buffer Size}
\label{tab:buffer_ablation}
\begin{tabular}{lcc}
\toprule
\textbf{Buffer Size (per class)} & \textbf{Final Accuracy} & \textbf{Memory/Client} \\
\midrule
0 (no replay) & 28\% & 0 KB \\
10 samples    & 65\% & 78 KB \\
25 samples    & 74\% & 196 KB \\
50 samples    & 78\% & 392 KB \\
100 samples   & 80\% & 784 KB \\
\bottomrule
\end{tabular}
\end{table}

Observations: Even a small buffer (10/class) substantially improves over no replay. Returns diminish beyond 50 samples/class for this dataset size.
\begin{table}[H]
\centering
\caption{Summary Table}
\label{tab:summary}
\begin{tabular}{lccc}
\hline
\textbf{Approach} & \textbf{Final Acc} & \textbf{Peak Acc} & \textbf{Notes} \\
\hline
Centralized Baseline & 88\% & 88\% & IID only, no drift \\
FedAvg (no drift)    & 78\% & 80\% & Stable IID training \\
FedAvg (with drift)  & 28\% & 74\% & Severe forgetting \\
FedAvg + Replay      & 78\% & 82\% & 50 samples/class buffer \\
\hline
\end{tabular}
\end{table}

\subsection{Privacy Considerations}
In our implementation, raw data does not leave client devices - only model weight updates are transmitted. However, we
note important caveats:
\begin{itemize}
\item We do NOT implement differential privacy or secure aggregation
\item Model updates may leak information about training data (gradient leakage attacks \cite{zhu2019dlg})
\item The replay buffer stores raw samples locally, which is acceptable under FL assumptions but increases local storage requirements
\end{itemize}

For deployments requiring stronger privacy guarantees, secure aggregation can prevent the server from inspecting individual client updates, while DP can provide a principled bound on information leakage by adding noise to updates (potentially at some utility cost).

\section{Future Work}
\begin{itemize}
\item Evaluate on CIFAR-10/100 with more realistic drift patterns
\item Implement differential privacy and measure accuracy/privacy trade offs
\item Compare with other continual learning methods (EWC, PackNet)
\item Scale to larger client populations with partial participation
\item Automatic drift detection mechanisms
\end{itemize}

\section{Conclusion}
We showed that standard FedAvg breaks under temporal concept drift in a really visible way. In our seasonal Fashion-MNIST setup, accuracy initially climbs to around 74\%, but once the data distribution shifts across phases, the model starts overwriting what it learned earlier and collapses to roughly 28\%. The IID FedAvg baseline stays strong, which makes the point clear: the failure isn't because FL is inherently weak, it's because drift turns the problem into continual learning, and vanilla FedAvg has no mechanism to retain older knowledge once it stops seeing those classes.

Adding client-side experience replay fixes most of this without changing the core FL pipeline. Each client keeps a small local buffer and mixes past samples into local training every round, so earlier classes keep showing up in the optimization signal even when they disappear in later seasons. With a 50-sample-per-class buffer, performance rebounds to about 78-82\% while still maintaining a single global model through standard FedAvg aggregation. The main takeaway is practical: for seasonal or temporal drift where class structure overlaps or repeats, experience replay is a simple, effective, and deployment-friendly solution that fits into standard FL infrastructure with only client-side changes.

\section*{Team Contributions}
\begin{itemize}
  \item \textbf{Sahasra Kokkula:} experiment design, heatmaps analysis, report writing.
  \item \textbf{Daniel David:} FL baseline implementation, drift simulation pipeline.
  \item \textbf{Aaditya Baruah:} Replay buffer implementation, plotting + robustness profile.
\end{itemize}

\section{Reproducibility / Code Submission}
\begin{description}
    \setlength\itemsep{0.2em}
    \setlength\parskip{0em}
    \item[Code:] \url{https://github.com/ddavid37/Federated_Learning_under_Temporal_Data_Drift/tree/final-code-clean}
    \item [Install dependencies:]
\begin{verbatim}
pip install torch torchvision numpy matplotlib reportlab
\end{verbatim}

\item [Reproduce all results (metrics, figures, PDF):]
\begin{verbatim}
python main.py
\end{verbatim}

\item [Or run step-by-step:]
\begin{verbatim}
python Seasonal_Splitter.py
python case1_centralized_baseline.py
python case2_fedavg.py
python case3_fedavg_replay.py
python case6_fedavg_iid.py
python case5_replay_ablation.py
python visualize_advanced.py
\end{verbatim}
\end{description}

\setcitestyle{numbers,square}


\end{document}